\documentclass[11pt,a4paper]{article}
\usepackage[hyperref]{acl2020}
\usepackage{times}
\usepackage{latexsym}
\usepackage{graphicx}
\usepackage{float}
\usepackage{array}

\usepackage{microtype}

\aclfinalcopy % Uncomment this line for the final submission
\title{Unsupervised Evaluation of Interactive Dialog with DialoGPT}

\author{Shikib Mehri and Maxine Eskenazi \\
  Dialog Research Center, Language Technologies Institute \\
  Carnegie Mellon University, USA \\
  \texttt{\{amehri,max\}@cs.cmu.edu}}

\date{}

\begin{document}
\maketitle
\begin{abstract}
It is important to define meaningful and interpretable automatic evaluation metrics for open-domain dialog research. Standard language generation metrics have been shown to be ineffective for dialog. This paper introduces the \textbf{FED} metric (fine-grained evaluation of dialog), an automatic evaluation metric which uses DialoGPT, without any fine-tuning or supervision. It also introduces the FED dataset which is constructed by annotating a set of human-system and human-human conversations with eighteen fine-grained dialog qualities. The FED metric (1) does not rely on a ground-truth response, (2) does not require training data and (3) measures fine-grained dialog qualities at both the turn and whole dialog levels. FED attains moderate to strong correlation with human judgement at both levels.

\end{abstract}

% intro: 1.5
% related work: 1 
% data collection: 1
% data analysis: 1
% methods: 1
% results: 2
% total: 7.5
\section{Introduction}

Evaluation metrics often define the research direction of a field. As dialog systems begin to demonstrate human-level performance, the development and adoption of meaningful and interpretable automatic evaluation measures is essential  \citep{zhang2019dialogpt,adiwardana2020towards}. Since standard metrics (e.g., BLEU, METEOR) have been shown to be ineffective for dialog \citep{deriu2019survey,liu2016not}, human evaluation is often used. However, it is typically only used as a final evaluation since it is costly. During development, systems are generally optimized for poorly correlated automatic metrics which can result in sub-par performance \citep{dinan2019second}. Automatic metrics must be meaningful and interpretable so that they can be used to compare dialog systems, understanding their respective strengths and weaknesses, and effectively guide dialog research.

Dialog evaluation is difficult for several reasons: (1) The one-to-many nature of dialog \citep{zhao2017learning} makes word-overlap metrics ineffective for scoring valid responses that deviate from the ground-truth \citep{liu2016not,gupta2019investigating}. (2) Dialog quality is inherently multi-faceted \citep{walker1997paradise,see2019makes} and an interpretable metric should measure several qualities (e.g., \textit{interesting}, \textit{relevant}, \textit{fluent}). (3) Dialog systems have begun to be evaluated in an interactive setting \citep{ram2018conversational,adiwardana2020towards} where a real user has a back-and-forth conversation with a system. Interactive evaluation is not constrained to a static corpus and better captures the performance of a system in a realistic setting. However, the existing automatic metrics compare to a ground-truth response, making them unsuitable for assessing interactive conversations. To address these three problems, this paper presents the \textbf{FED} metric (fine-grained evaluation of dialog) which assesses eighteen qualities of dialog without relying on a reference response. 

First, a dataset of human quality annotations is collected for the human-system (Meena and Mitsuku) and human-human conversations released by \citet{adiwardana2020towards}. Dialogs are annotated at both the turn level and the dialog level for eighteen fine-grained dialog qualities. This FED dataset can be used to benchmark the performance of automatic metrics relative to human judgement. Analysis of this data provides insight into the qualities of dialog that are most important to human annotators. It therefore highlights the qualities that should be the focus of attention in dialog research.

The FED dataset is intended only for evaluating automatic metrics relative to human judgement. It does not consist of any training data. As such, this paper addresses the task of developing an automatic evaluation metric which (1) does not compare to a reference response, (2) assesses eighteen different qualities of dialog and (3) relies on no training data or supervision. This paper is the first, to the best of our knowledge, to address this important and challenging problem. 

The FED metric described here leverages a massively pre-trained model, DialoGPT \citep{zhang2019dialogpt}, which can generate practically human-level responses. \citet{kocijan2019surprisingly} assert that pre-trained models implicitly capture world knowledge and can therefore perform common-sense reasoning. Similarly, we posit that DialoGPT has implicitly captured some notion of dialog quality and can therefore be used for dialog evaluation. \citet{eskenazi2019beyond} assessed the quality of a system utterance in an interactive setting by looking at the \textit{following user response}. The proposed evaluation metric is based on the same intuition. Given a system response, its quality is measured by computing the likelihood that DialoGPT will respond to it with a particular follow-up utterance (e.g., \textit{``That is really interesting!''}). DialoGPT is more likely to respond in this way to what it believes is an \textit{interesting} system response. A set of follow-up utterances is constructed for each of the eighteen qualities and the likelihoods of these follow-up utterances are used to measure dialog quality. 

The FED metric obtains moderate to strong correlation with human judgement for turn-level and dialog-level evaluation without any training data or ground-truth response. Analysis in this paper demonstrates that through large-scale pre-training, DialoGPT has implicitly captured some notion of dialog quality. These results suggest that pre-trained models can be leveraged to further improve dialog evaluation. 

The contributions of this paper are as follows: (1) The FED dataset\footnote{\url{http://shikib.com/fed_data.json}} was collected for fine-grained evaluation of interactive dialog, with annotations for eighteen dialog qualities at both the turn- and the dialog-level. (2) Analysis of the FED dataset identifies the dialog qualities most important to human annotators. (3) DialoGPT is shown to implicitly capture an understanding of dialog quality. (4) The FED metric\footnote{\url{https://github.com/shikib/fed}} has moderate to strong correlation with human judgement by leveraging DialoGPT, without training data or reference responses.

\section{Related Work}

\subsection{Automatic Dialog Evaluation}

Standard automatic metrics for language generation have been shown to correlate poorly with human judgement of dialog \citep{liu2016not,lowe-etal-2017-towards,gupta2019investigating}. This poor performance can largely be explained by the one-to-many nature of dialog \citep{zhao2017learning}. To avoid comparing to a single reference response, several authors have proposed using multiple reference responses. Multiple reference responses can be obtained with retrieval models \citep{galley-etal-2015-deltableu,Sordoni2015ANN} or through data collection \citep{gupta2019investigating}. These multi-reference metrics show performance improvement, but it is infeasible to thoroughly cover the space of all potential responses. The FED metric does not rely on a ground-truth response.

\citet{lowe-etal-2017-towards} train ADEM to produce a quality score conditioned on the dialog context, the reference response and the generated response. \citet{venkatesh2018evaluating} present a framework for evaluating Alexa prize conversations which attains moderate correlation with user ratings. Both methods are trained on explicit quality annotations. In contrast, the FED metric proposed here requires no supervision.

\citet{mehri2020usr} introduce USR, an unsupervised and reference-free evaluation metric for dialog generation. Similar to FED, USR uses pre-trained models to assess several dialog qualities. However, they are limited to five qualities with hand-designed models and unsupervised tasks for each quality. In comparison, FED is more general and encapsulates eighteen dialog qualities.

\subsection{Dialog Qualities}

Human evaluation in dialog is often limited to only measuring overall quality or response appropriateness. However, dialog quality is multi-faceted and should not be reduced to a single measurement.

PARADISE \citep{walker1997paradise}, one of the first frameworks for dialog evaluation, measured several different properties of dialog and combined them to estimate user satisfaction. \citet{see2019makes} used a variety of human judgements for dialog including interestingness, making sense, avoiding repetition, fluency, listening and inquisitiveness. \citet{see2019makes} emphasize the importance of measuring multiple qualities when evaluating dialog systems. There are several examples of human evaluation of multiple dialog qualities. \citet{gopalakrishnan2019topical} annotate system responses using: interesting, comprehensible, on-topic and use of knowledge. \citet{shin2019happybot} measure empathy, fluency and relevance. \citet{zhang2019dialogpt} evaluate responses using relevance, informativeness and human-likeness. \citet{adiwardana2020towards} evaluate in both static and interactive environments using specificity and sensibleness.

\begin{table*}
    \centering
    \begin{tabular}{|m{0.6\linewidth}|m{0.3\linewidth}|}
    \hline
        \textbf{Question} &  \textbf{Used By} \\ \hline
        To the average person, is the response \textbf{interesting}? & \citet{see2019makes,gopalakrishnan2019topical,mehri2020usr}\\ \hline
        Is the response \textbf{engaging}? & \citet{yi2019towards} \\ \hline
        Is the response \textbf{generic} or \textbf{specific} to the conversation? & \citet{adiwardana2020towards} \\ \hline
        Is the response \textbf{relevant} to the conversation? & \citet{see2019makes,gopalakrishnan2019topical,shin2019happybot,zhang2019dialogpt,mehri2020usr} \\ \hline
        Is the response \textbf{correct} or was there a misunderstanding of the conversation? & None specifically \\  \hline
        Is the response \textbf{semantically appropriate}? & \citet{see2019makes} \\ \hline
        Is the response \textbf{understandable}? & \citet{gopalakrishnan2019topical,mehri2020usr} \\ \hline 
        Is the response \textbf{fluently written}? & \citet{see2019makes,shin2019happybot,zhang2019dialogpt,ghandeharioun2019approximating,mehri2020usr} \\ \hline
        \textbf{Overall impression} of the response? & Many \\ \hline
        
    \end{tabular}
    \caption{The questions asked for turn-level annotation. Examples of prior work that has used each dialog quality are listed. No one has specifically used \textit{Correct}, however its meaning is often encapsulated in \textit{Relevant}.}
    \label{tab:turnlevelqualities}
\end{table*}

\subsection{Pre-trained Dialog Models}

The success of pre-trained language models \citep{radford2018improving,devlin2018bert} has recently been extended to the domain of dialog. \citet{zhang2019dialogpt} pre-train DialoGPT on Reddit and attain human-level performance on the task of response generation. The open-source DialoGPT model was used to construct the FED metric presented in this paper. \citep{adiwardana2020towards} similarly pre-trained their Meena dialog system on an unspecified large conversational dataset.

\section{Data Collection}

A dataset of human quality annotations was collected to assess automatic metrics by measuring correlation with human judgements. \citet{adiwardana2020towards} collected a set of conversations\footnote{https://github.com/google-research/google-research/tree/master/meena} between a human and two open-domain dialog systems, Meena \citep{adiwardana2020towards} and Mitsuku\footnote{https://medium.com/pandorabots-blog/mitsuku-wins-loebner-prize-2018-3e8d98c5f2a7}. In addition, they also released human-human dialogs collected in the same environment where one of the humans was selected to play the role of the system. We annotated a subset of these conversations with human quality judgements to create the FED dataset.

Workers on Amazon Mechanical Turk (AMT) annotated 40 Human-Meena conversations, 44 Human-Mitsuku conversations and 40 Human-Human conversations. For each conversation, three system responses were hand-selected to be annotated at the turn level, presented to the worker sequentially. Then the worker was shown the entire conversation and annotated on the dialog level. Five workers annotated each conversation. They did not know which system was involved in a conversation, since all mentions of the system name were replaced with the word ``System.''

Since dialog quality is inherently multi-faceted it is important to measure several different qualities of dialog. Eighteen fine-grained dialog qualities are measured in the FED dataset: eight at the turn level and ten at the dialog level.

\subsection{Turn-Level Annotation}

Given a dialog context and a system response, the worker assessed the response according to eight fine-grained measures as well as for overall quality. The list of turn-level measures is shown in Table \ref{tab:turnlevelqualities}. The options for each of the fine-grained qualities were: \textit{No}, \textit{Somewhat}, \textit{Yes}, \textit{N/A}. For \textit{understandable}, the \textit{Somewhat} option was not provided, similar to prior past work \citep{gopalakrishnan2019topical}. Responding \textit{N/A} required written justification. The overall impression question was measured on a five-point Likert scale.

The workers were given detailed instructions and examples for each  question presented in Table \ref{tab:turnlevelqualities}. These instructions are provided in the supplementary materials.

\subsection{Dialog-Level Annotation}
For dialog-level annotation, workers were asked to label the quality of a system over the duration of an entire conversation. The dialog-level questions listed in Table \ref{tab:dialoglevelqualities} cover ten fine-grained dialog qualities and an additional question on overall impression. The available options for each of the fine-grained qualities were \textit{No}, \textit{Somewhat}, \textit{Yes}, \textit{N/A}. For \textit{consistency}, the \textit{Somewhat} option was not provided because the existence of an inconsistency is binary. Overall impression was measured on a five-point Likert scale.

\begin{table*}
    \centering
    \begin{tabular}{|m{0.6\linewidth}|m{0.3\linewidth}|}
    \hline
        \textbf{Question} &  \textbf{Used By} \\ \hline
        Throughout the dialog, is the system \textbf{coherent} and maintain a good conversation flow? & \citet{see2019makes} \\ \hline
        Is the system able to \textbf{recover from errors} that it makes? & None \\ \hline
        Is the system \textbf{consistent} in the information it provides throughout the conversation? & \citet{qin2019entity} \\ \hline
        Is there \textbf{diversity} in the system responses? & \citet{see2019makes,ghandeharioun2019approximating} \\ \hline
        Does the system discuss topics in \textbf{depth}? & \citet{guo2018topic} \\ \hline
        Does the system display a \textbf{likeable} personality? & \citet{shin2019happybot,ghandeharioun2019approximating} \\ \hline
        Does the system seem to \textbf{understand} the user? & \citet{see2019makes} \\ \hline 
        Is the system \textbf{flexible and adaptable} to the user and their interests? & \citet{guo2018topic} \\ \hline
        Is the system \textbf{informative} throughout the conversation? & \citet{zhang2019dialogpt} \\ \hline
        Is the system \textbf{inquisitive} throughout the conversation?  & \citet{see2019makes} \\ \hline
        \textbf{Overall impression} of the dialog? & Many \\ \hline
        
    \end{tabular}
    \caption{The qualities annotated at the dialog-level. Examples of prior work that has used each dialog quality are listed. To our knowledge, error recovery has not been used for human evaluation.}
    \label{tab:dialoglevelqualities}
\end{table*}

\subsection{Dataset Statistics}

A total of 124 conversations were annotated (40 Meena, 44 Mitsuku, 40 Human). Five different workers saw each conversation (HIT). Each conversation had one dialog-level annotation and three turn-level annotations for chosen system responses that were randomly sampled from the conversation. There were 9 questions for turn-level annotation and 11 for dialog-level annotation. In total, the FED dataset includes 3348 turn-level and 1364 dialog-level data points, for a total of 4712. This dataset intended to be used solely for the evaluation of metrics, as the number of annotated conversations is not large enough to accommodate both training and testing.

\subsection{Data Processing}

Given that each of the 4712 data points was labeled by five annotators, post-processing was used to improve the quality of the data through the removal of outliers. Given five annotations for a given question, the furthest label from the mean is removed if its distance from the mean is greater than half the standard deviation of the five annotations.

\section{Data Analysis}

The fine-grained nature of the FED dataset is grounds for a rich analysis. First, inter-annotator agreement is evaluated for all of the dialog qualities. Next, the dataset is used to better understand the comparative strengths and weaknesses of the three systems (Mitsuku, Meena, Human). Finally, detailed analysis of the data provides insight into the fine-grained qualities that most strongly contribute to the annotators' overall impression.

\subsection{Inter-Annotator Agreement}

To compute inter-annotator agreement, the correlation between each annotation and the mean of the five (or four, after outlier removal) annotations for the same question is measured. The Spearman correlation for each turn-level and dialog-level question is shown in Table \ref{tab:agreement}

\begin{table}
    \centering
    \begin{tabular}{|c|c|} \hline
        \textbf{Quality} & \textbf{Spearman}  \\ \hline
\multicolumn{2}{|c|}{Turn-Level} \\ \hline
Interesting & 0.819\\
Engaging & 0.798\\
Specific & 0.790\\
Relevant & 0.753\\
Correct & 0.780\\
Semantically Appropriate & 0.682\\
Understandable & 0.522\\
Fluent & 0.714 \\
Overall Impression & 0.820 \\ \hline
\multicolumn{2}{|c|}{Dialog-Level} \\ \hline
Coherent & 0.809 \\
Error Recovery & 0.840 \\
Consistent & 0.562 \\
Diverse & 0.789 \\
Topic Depth & 0.833 \\
Likeable & 0.838 \\
Understanding & 0.809 \\
Flexible & 0.816 \\
Informative & 0.806 \\
Inquisitive & 0.769 \\
Overall Impression & 0.830    \\ \hline
\end{tabular}
    \caption{Spearman correlation for each of the dialog qualities. The correlation was measured by correlating each annotation with the mean of the other annotations for the same question.}
    \label{tab:agreement}
\end{table}

Inter-annotator agreement is high for all of the dialog qualities, suggesting that all of the qualities were well-understood by the annotators and relevant and that the instructions removed much of the ambiguity from the task. Two qualities, \textit{understandable} and \textit{consistent}, have slightly lower correlations, in the 0.5 - 0.6 range. These qualities did not include \textit{Somewhat} as an answer. This probably contributed to the lower inter-annotator agreement.

\begin{table}
    \centering
    \begin{tabular}{|m{0.28\linewidth}|c|c|c|} \hline
        \textbf{Quality} & \textbf{Mitsuku} & \textbf{Meena} & \textbf{Human}  \\ \hline
\multicolumn{4}{|c|}{Turn-Level} \\ \hline
Interesting & 2.30 & \textbf{2.58} & 2.35 \\
Engaging & 2.53 & \textbf{2.75} & 2.49 \\
Specific & 2.48 & \textbf{2.74} & 2.56\\
Relevant & 2.80 & \textbf{2.88} & 2.74\\
Correct & 2.74 & \textbf{2.84} & 2.66\\
Semantically- Appropriate & 2.84 & \textbf{2.92} & 2.85\\
Understandable & \textbf{0.97} & \textbf{0.97} & 0.94\\
Fluent & 2.83 & \textbf{2.90} & 2.80 \\
Overall  & 3.81 & \textbf{4.19} & 3.85 \\ \hline
\multicolumn{4}{|c|}{Dialog-Level} \\ \hline
Coherent & 2.20 & 2.88 & \textbf{2.94} \\
Error Recovery & 2.22 & 2.69 & \textbf{2.86} \\
Consistent & 0.82 & 0.95 & \textbf{0.98} \\
Diverse & 2.23 & 2.46 & \textbf{2.88} \\
Topic Depth & 1.80 & 2.28 & \textbf{2.78} \\
Likeable & 2.10 & 2.61 & \textbf{2.97} \\
Understanding & 2.23 & 2.86 & \textbf{2.98} \\
Flexible & 2.22 & 2.72 & \textbf{2.97} \\
Informative & 2.10 & 2.60 & \textbf{2.85} \\
Inquisitive & 2.35 & 2.76 & \textbf{2.88} \\
Overall  & 3.10 & 4.11 & \textbf{4.60}    \\ \hline
\end{tabular}
    \caption{Performance of each system on the fine-grained qualities. All scores are 1-3, except Understandable and Consistent are 0-1 and Overall is 1-5.}
    \label{tab:systemscores}
\end{table}
\subsection{System Performance}

While \citet{adiwardana2020towards} presented a performance comparison between Mitsuku, Meena and Humans in an interactive setting, their evaluation only used two qualities: \textit{specificity} and \textit{sensibility}. In contrast, the FED dataset has eighteen fine-grained qualities thus providing more information about the strengths and weaknesses of each system.

The fine-grained performance of each system shown in Table \ref{tab:systemscores}. For all of the turn-level qualities, Meena outperforms both Mitsuku and Human. The strength of Meena is most noticeable for \textit{interesting}, \textit{engaging} and \textit{specific}.

However, turn-level qualities are insufficient to evaluate a dialog system. Dialog is by definition a multi-turn interaction. Thus, in some cases, a sub-optimal system response might result in a better long-term dialog. Humans significantly outperform the two systems for dialog-level qualities. The difference between Meena and Mitsuku is very pronounced at the dialog level, with a 1 point difference in overall score. The higher variance in scores and the stronger performance of human dialogs, shows that dialog-level evaluation is reliable than turn-level. Meena's scores suggest that it is fairly \textit{coherent}, \textit{understanding} and \textit{flexible}. However, it struggles with \textit{diversity}, \textit{topic depth} and \textit{likeable}.

\subsection{Fine-Grained Quality Analysis}

The FED dataset can be used to examine the relative importance of each fine-grained dialog quality by measuring its contribution to the overall impression. For both turn-level and dialog-level, a regression is trained to predict the overall score given the fine-grained qualities as input. The regression weights provide insight into the fine-grained qualities that most contribute to the overall impression as labeled by human annotators. A softmax is computed over the regression weights to determine the relative contribution of each fine-grained dialog quality. A dialog quality with a higher weight contributes more to the human's overall impression. The results are shown in Table \ref{tab:importance}.

\begin{table}
    \centering
    \begin{tabular}{|c|c|} \hline
        \textbf{Quality} & \textbf{Importance (\%)}  \\ \hline
\multicolumn{2}{|c|}{Turn-Level} \\ \hline
\textbf{Interesting} & \textbf{16.15}\\
Engaging & 7.46\\
Specific & 9.64\\
\textbf{Relevant} & \textbf{18.10}\\
Correct & 13.77\\
Semantically Appropriate & 9.90\\
Understandable & 10.70\\
\textbf{Fluent} & \textbf{14.27} \\ \hline
\multicolumn{2}{|c|}{Dialog-Level} \\ \hline
\textbf{Coherent} & \textbf{10.95} \\
Error Recovery & 9.15 \\
Consistent & 7.92 \\
Diverse & 10.09 \\
Topic Depth & 10.51 \\
\textbf{Likeable} & \textbf{12.03} \\
\textbf{Understanding} & \textbf{11.01} \\
Flexible & 10.34 \\
Informative & 8.00 \\
Inquisitive & 9.50 \\ \hline
\end{tabular}
    \caption{Relative importance of each dialog quality for predicting the overall impression. The most important qualities for turn-level and dialog-level are in bold.}
    \label{tab:importance}
\end{table}

The most important turn-level qualities are \textit{interesting}, \textit{relevant} and \textit{fluent}. This suggests that developing a system that is consistently interesting, relevant and fluent will result in the highest improvement in the user's overall impression. There is less variance in the importance of dialog-level qualities than in the turn-level qualities possibly because there is less overlap in meaning amongst the qualities and all of the dialog-level qualities seem somewhat important. The most important dialog-level qualities are \textit{coherent}, \textit{likeable} and \textit{understanding}. Improving a system's coherence, understanding of the user and its likeableness would thus be the most likely way to improve the overall impression of a dialog system. 

\section{Methods}

The FED (fine-grained evaluation of dialog) metric is an automatic evaluation metric for dialog which (1) does not need to compare to a reference response, (2) measures eighteen fine-grained qualities of dialog, and (3) does not use training data. Capturing a diverse set of fine-grained qualities without supervision is an especially challenging problem.

The development of the FED metric is motivated by two areas of prior work: (1) pre-trained language models and their capabilities and (2) the use of follow-up utterances as a means of evaluation.

\subsection{DialoGPT}

\citet{zhang2019dialogpt} extend GPT-2 \citep{radford2018improving} to train DialoGPT on 147M conversation-like interactions from Reddit. As per their evaluation, DialoGPT outperforms humans at producing relevant, interesting and human-like responses. 

\citet{kocijan2019surprisingly} show that pre-trained language models, specifically BERT \citep{devlin2018bert}, implicitly capture world knowledge and can therefore perform common sense reasoning. By calculating which answer results in a more probable sentence according to BERT, they strongly outperform other methods on the Winograd Schema Challenge \citep{levesque2012winograd}.

Just as BERT has been shown to capture world knowledge, we posit that DialoGPT has implicitly captured some notion of dialog quality. The qualities of a particular dialog context (e.g., \textit{interesting}, \textit{relevant}, \textit{informative}) likely inform DialoGPT's response and, as such, must be captured by the model. If there was training data for the eighteen dialog qualities, this hypothesis could be verified by fine-tuning DialoGPT for the task of dialog evaluation. Without training data, however, the challenge is to devise an unsupervised mechanism for extracting the quality information captured by DialoGPT.

\subsection{Follow-Up Utterance for Evaluation}

\citet{eskenazi2019beyond} assess the quality of a system utterance in an interactive setting, by looking at the \textit{following user response}. When users speak to a system, their response to a given system utterance may implicitly or explicitly provide feedback for the system. For example, if a user follows up a system utterance with \textit{``That's not very interesting''}, they are providing information about the quality of the system utterance. 

The conversations in the FED dataset were collected in an interactive setting. Thus the use of the follow-up utterance is a valid option. Even if users consistently provided feedback, it would be difficult to interpret without training data.

\subsection{Evaluating with DialoGPT}

The proposed FED metric is motivated by (1) the intuition that DialoGPT has implicitly learned to reveal dialog quality and (2) that the follow-up utterance can provide valuable information about a system response. To measure the quality of a system response $s$, we compute the likelihood of the model generating various follow-up utterances (e.g., \textit{``Wow! Very interesting.''}) in response to $s$. DialoGPT will be more likely to respond with a positive follow-up utterance if given a better (e.g., more \textit{interesting}/\textit{relevant}/\textit{fluent}) preceding system utterance.

For each of the eighteen fine-grained dialog qualities, a set of positive follow-up utterances, $p$, and a set of negative follow-up utterances, $n$, is constructed. Specifically, given a dialog context $c$, a system response $r$ and a function $\mathcal{D}$ that computes the log-likelihood of DialoGPT generating a particular response, the predicted score for a dialog quality is calculated as:

\begin{equation}
    \sum_{i=1}^{|p|} \mathcal{D}(c + r, p_i) - \sum_{i=1}^{|n|} \mathcal{D}(c + r, n_i)
    \label{modeleq}
\end{equation}

This equation can be modified to predict scores for dialog-level qualities, by simply removing the system response $r$ from the equation. 

A response is said to be \textit{interesting} if it is more likely that DialoGPT (acting as the user) responds with a positive follow-up utterance (e.g., \textit{``Wow! Very interesting''}) than with a negative one (e.g., \textit{``That's really boring''}). For each of the eighteen qualities, several positive and negative utterances were hand-written and minimally tuned on a small subset of the dataset (10 conversations). Follow-up utterances for each quality are provided in the supplementary materials.

Generally, negative follow-up utterances are more meaningful than positive ones. For example, if a system response is \textit{irrelevant}, a follow-up utterance of \textit{``That's not relevant''} is reasonable. However, acknowledging the relevance of a system response is less likely. Therefore the log-likelihood produced by DialoGPT will be noisier and less informative. The number of positive utterances for each dialog quality ranges between 0 and 4, and the number of negative utterances ranges between 1 and 4. While the fine-grained qualities are computed in this manner, the overall impression scores are calculated as an average of the scores for either the turn-level or dialog-level qualities. 

\section{Results}

\subsection{Experimental Setup}

The FED metric was evaluated using four variations of the pre-trained DialoGPT model. The pre-trained DialoGPT models can be either medium size: 345M or large: 762M. They are either fine-tuned from GPT-2 \citep{radford2018improving} or trained from scratch. 
The follow-up utterances were handwritten and minimally tuned on 10 conversations using the 762M fine-tuned model. The small (117M) DialoGPT model was not used since \citet{zhang2019dialogpt} demonstrated its poor performance.

Most of the turn-level qualities were scored using only the last system response as context. For \textit{relevant}, \textit{correct} and dialog-level metrics, the entire conversation was used as context. 

\subsection{Correlation with Human Judgement}

The Spearman correlation was measured between the predicted quality scores and the mean of the annotated scores. Correlations for all the dialog qualities, and all four variations of the underlying DialoGPT model are shown in Table \ref{tab:metriccorrelation}.

\begin{table*}
    \centering
    \begin{tabular}{|l|cccc|} \hline
        \textbf{Quality} & \textbf{345M fs} & \textbf{345M ft} & \textbf{762M fs} & \textbf{762M ft}  \\ \hline
\multicolumn{5}{|c|}{Turn-Level} \\ \hline
Interesting & 0.388 & \textbf{0.431} & 0.406 & 0.408 \\
Engaging &  0.268 & 0.285 & 0.278 & \textbf{0.318} \\
Specific & 0.260 & \textbf{0.326} & 0.270 & 0.267 \\
Relevant & \textit{0.028} & \textit{-0.027} & \textit{0.001} & \textbf{0.152} \\
Correct &  \textit{0.000} & \textit{0.037} & \textit{0.020} & \textbf{0.133} \\
Semantically Appropriate  & \textit{0.040} & \textbf{0.177} & 0.141 & 0.155 \\
Understandable & \textit{0.047} & \textit{0.048} & \textit{0.075} & \textbf{0.111} \\
Fluent &  0.157 & 0.184  & 0.133 & \textbf{0.224} \\
Overall  &  0.122  & \textit{0.092} & \textit{0.094} & \textbf{0.209} \\ \hline
\multicolumn{5}{|c|}{Dialog-Level} \\ \hline
Coherent &  0.195 & \textit{0.151} & \textit{0.149} & \textbf{0.251} \\
Error Recovery &  \textit{0.165} & \textit{0.128} & \textit{0.126} & \textit{0.165} \\
Consistent &  \textit{0.041} & \textit{0.011} & \textit{0.006} & \textit{0.116} \\
Diverse &  \textbf{0.449} & 0.431 & 0.414 & 0.420 \\
Topic Depth &  \textbf{0.522} & 0.479 & 0.470 & 0.476 \\
Likeable &  \textit{0.047} & \textit{0.172} & 0.224 & \textbf{0.262} \\
Understanding &  0.237 & 0.174 & 0.192 & \textbf{0.306} \\
Flexible &  0.260 & \textbf{0.408} & 0.298 & 0.293 \\
Informative &  0.264 & 0.328 & \textbf{0.337} & 0.288 \\
Inquisitive & \textit{0.137} & \textit{0.143} & \textbf{0.298} & 0.163 \\
Overall  &  0.401 & 0.359 & 0.355 & \textbf{0.443} \\\hline
\end{tabular}
    \caption{Spearman correlations with human judgement. All values that are not statistically significant ($p > 0.05$) are italicized. The highest correlation for each quality is shown in bold.} 
    \label{tab:metriccorrelation}
\end{table*}

The best overall turn-level correlation is \textbf{0.209} and the best overall dialog-level correlation is \textbf{0.443}. To our knowledge, there are presently no other metrics that operate without a ground-truth response, thus these results cannot be directly compared to any existing metrics. However, prior work on dialog evaluation reveals roughly similar correlation. Multi-reference evaluation for dialog achieves correlations in the 0.10 - 0.27 range \citep{gupta2019investigating} and ADEM has correlations in the 0.28 - 0.42 range \citep{lowe-etal-2017-towards}. Given neither training data nor ground-truth response, the FED metric performs competitively relative to this prior work.

\subsection{Discussion}

The FED metric works better for some dialog qualities than others. This is because DialoGPT was trained on Reddit. It is more likely that it has captured certain dialog qualities that Reddit exhibits. For example, it is more likely that DialoGPT learns to measure qualities like \textit{interesting} and \textit{engaging}, than \textit{understandable} and \textit{consistent}. In the Reddit training data, the former two qualities show more variation than the latter. For example, there are interesting and un-interesting utterances, however most utterances on Reddit are generally understandable. The former two qualities are also more likely to influence the system response. Conversely, the latter two qualities are unlikely to be acknowledged in the response. For example, since Reddit is a multi-participant forum and not a one-on-one conversation, inconsistencies in conversation history are unlikely to be reflected in the response. As such, it is unsurprising that this approach struggles to measure the consistency of a dialog. 

An optimal generation model (e.g., a human) should exhibit compositionality and be capable of producing utterances that have never been observed. For example, even if \textit{`That is not consistent'} has never appeared in the training data, a compositional model would be capable of generating it. This difference in performance across the different dialog qualities suggests that DialoGPT exhibits some degree of compositionality, as evidenced by its ability to compose some follow-up utterances which are not frequently observed in the Reddit data (e.g., \textit{`You really don't know much?'}), however it still struggles with follow-up utterances consisting of less frequently observed concepts (e.g., \textit{consistent}, \textit{understandable}).

DialoGPT could be used to better measure these qualities by fine-tuning on additional conversational data from a source other than Reddit or on a training set annotated with human quality judgements. However, even without additional fine-tuning, FED effectively measures many qualities.

This paper has carried out an assessment of the FED metric for three open-domain conversation agents: Meena, Mitsuku and Human. Since these three systems are different in nature and FED exhibits strong correlation with human judgements across all the systems, we believe that the performance of FED will hold for other open-domain dialog systems and will not be restricted to a particular type of model or a specific dataset. However, the FED dataset consists of only open-domain chit-chat conversations. As such, future work is needed to determine whether the FED metric will generalize to goal-oriented dialog. Since DialoGPT has not observed goal-oriented training data, it may be necessary to use self-supervised fine-tuning on the new domain \citep{mehri2020usr}.

As with all automated metrics, there is the potential to game the FED metric and obtain artificially high scores, especially by having a model produce responses that are likely to result in specific follow-up utterances. To this end, the FED metric is not a replacement for human evaluation. It is instead a means of measuring dialog quality for the purposes of validation and model tuning.

The FED metric is (1) unsupervised, (2) does not rely on a reference response and (3) can be used to assess many dialog qualities. By having DialoGPT play the role of the user and assign probabilities to follow-up utterances, we have devised a mechanism of extracting information about dialog quality without any supervision. This mechanism is versatile and could potentially be extended to other dialog qualities. 

\section{Conclusion}

This paper introduces the FED dataset and the FED metric. The FED dataset is constructed by annotating a set of interactive conversations with eighteen fine-grained dialog qualities. The FED metric can be used to measure fine-grained qualities of dialog without comparing to a ground-truth response. By having DialoGPT take the role of the user and calculate the likelihood of follow-up utterances, the FED metric attains moderate to strong correlation with human judgement, without the use of any training data. The FED metric is inherently versatile and generalizable, making it applicable to other dialog qualities, domains or tasks. Both the FED dataset and the code for the FED metric will be released upon acceptance of this paper.

This paper sets the groundwork for several directions of future work. (1) The FED dataset can be used to benchmark automatic evaluation metrics on eighteen fine-grained dialog qualities. (2) Building on this paper, future work could identify mechanisms that further leverage pre-trained models for dialog evaluation. (3) Future work can explore strategies for extending the FED metric beyond open-domain chit-chat conversations to goal oriented dialog. (4) The FED metric can be used to evaluate, analyze and improve dialog systems.

%By effectively leveraging DialoGPT, the FED metric measures fine-grained dialog qualities without the use of a ground-truth response or any training data. The FED metric is shown to attain moderate to strong correlation with human judgement, demonstrating that dialog can be evaluated without a ground-truth response or training data. 

%The FED metric attains moderate to strong correlation with human judgement, demonstrating that DialoGPT has implicitly captured some notion of dialog quality. Future work can (1) use the FED dataset to benchmark automatic metrics, (2) better leverage pre-trained models for dialog evaluation and (3) devise better pre-trained models specifically for dialog evaluation. %to mitigate the shortcomings of DialoGPT on certain dialog qualities. 

\bibliography{acl2020}

\begin{thebibliography}{26}
\expandafter\ifx\csname natexlab\endcsname\relax\def\natexlab#1{#1}\fi

\bibitem[{Adiwardana et~al.(2020)Adiwardana, Luong, So, Hall, Fiedel,
  Thoppilan, Yang, Kulshreshtha, Nemade, Lu et~al.}]{adiwardana2020towards}
Daniel Adiwardana, Minh-Thang Luong, David~R So, Jamie Hall, Noah Fiedel, Romal
  Thoppilan, Zi~Yang, Apoorv Kulshreshtha, Gaurav Nemade, Yifeng Lu, et~al.
  2020.
\newblock Towards a human-like open-domain chatbot.
\newblock \emph{arXiv preprint arXiv:2001.09977}.

\bibitem[{Deriu et~al.(2019)Deriu, Rodrigo, Otegi, Echegoyen, Rosset, Agirre,
  and Cieliebak}]{deriu2019survey}
Jan Deriu, Alvaro Rodrigo, Arantxa Otegi, Guillermo Echegoyen, Sophie Rosset,
  Eneko Agirre, and Mark Cieliebak. 2019.
\newblock Survey on evaluation methods for dialogue systems.
\newblock \emph{arXiv preprint arXiv:1905.04071}.

\bibitem[{Devlin et~al.(2018)Devlin, Chang, Lee, and
  Toutanova}]{devlin2018bert}
Jacob Devlin, Ming-Wei Chang, Kenton Lee, and Kristina Toutanova. 2018.
\newblock Bert: Pre-training of deep bidirectional transformers for language
  understanding.
\newblock \emph{arXiv preprint arXiv:1810.04805}.

\bibitem[{Dinan et~al.(2019)Dinan, Logacheva, Malykh, Miller, Shuster, Urbanek,
  Kiela, Szlam, Serban, Lowe et~al.}]{dinan2019second}
Emily Dinan, Varvara Logacheva, Valentin Malykh, Alexander Miller, Kurt
  Shuster, Jack Urbanek, Douwe Kiela, Arthur Szlam, Iulian Serban, Ryan Lowe,
  et~al. 2019.
\newblock The second conversational intelligence challenge (convai2).
\newblock \emph{arXiv preprint arXiv:1902.00098}.

\bibitem[{Eskenazi et~al.(2019)Eskenazi, Mehri, Razumovskaia, and
  Zhao}]{eskenazi2019beyond}
Maxine Eskenazi, Shikib Mehri, Evgeniia Razumovskaia, and Tiancheng Zhao. 2019.
\newblock Beyond turing: Intelligent agents centered on the user.
\newblock \emph{arXiv preprint arXiv:1901.06613}.

\bibitem[{Galley et~al.(2015)Galley, Brockett, Sordoni, Ji, Auli, Quirk,
  Mitchell, Gao, and Dolan}]{galley-etal-2015-deltableu}
Michel Galley, Chris Brockett, Alessandro Sordoni, Yangfeng Ji, Michael Auli,
  Chris Quirk, Margaret Mitchell, Jianfeng Gao, and Bill Dolan. 2015.
\newblock delta{BLEU}: A discriminative metric for generation tasks with
  intrinsically diverse targets.
\newblock In \emph{Proceedings of the 53rd Annual Meeting of the Association
  for Computational Linguistics and the 7th International Joint Conference on
  Natural Language Processing (Volume 2: Short Papers)}, pages 445--450,
  Beijing, China. Association for Computational Linguistics.

\bibitem[{Ghandeharioun et~al.(2019)Ghandeharioun, Shen, Jaques, Ferguson,
  Jones, Lapedriza, and Picard}]{ghandeharioun2019approximating}
Asma Ghandeharioun, Judy~Hanwen Shen, Natasha Jaques, Craig Ferguson, Noah
  Jones, Agata Lapedriza, and Rosalind Picard. 2019.
\newblock Approximating interactive human evaluation with self-play for
  open-domain dialog systems.
\newblock In \emph{Advances in Neural Information Processing Systems}, pages
  13658--13669.

\bibitem[{Gopalakrishnan et~al.(2019)Gopalakrishnan, Hedayatnia, Chen,
  Gottardi, Kwatra, Venkatesh, Gabriel, Hakkani-T{\"u}r, and
  AI}]{gopalakrishnan2019topical}
Karthik Gopalakrishnan, Behnam Hedayatnia, Qinlang Chen, Anna Gottardi, Sanjeev
  Kwatra, Anu Venkatesh, Raefer Gabriel, Dilek Hakkani-T{\"u}r, and
  Amazon~Alexa AI. 2019.
\newblock Topical-chat: Towards knowledge-grounded open-domain conversations.
\newblock \emph{Proc. Interspeech 2019}, pages 1891--1895.

\bibitem[{Guo et~al.(2018)Guo, Metallinou, Khatri, Raju, Venkatesh, and
  Ram}]{guo2018topic}
Fenfei Guo, Angeliki Metallinou, Chandra Khatri, Anirudh Raju, Anu Venkatesh,
  and Ashwin Ram. 2018.
\newblock Topic-based evaluation for conversational bots.
\newblock \emph{arXiv preprint arXiv:1801.03622}.

\bibitem[{Gupta et~al.(2019)Gupta, Mehri, Zhao, Pavel, Eskenazi, and
  Bigham}]{gupta2019investigating}
Prakhar Gupta, Shikib Mehri, Tiancheng Zhao, Amy Pavel, Maxine Eskenazi, and
  Jeffrey~P Bigham. 2019.
\newblock Investigating evaluation of open-domain dialogue systems with human
  generated multiple references.
\newblock \emph{arXiv preprint arXiv:1907.10568}.

\bibitem[{Kocijan et~al.(2019)Kocijan, Cretu, Camburu, Yordanov, and
  Lukasiewicz}]{kocijan2019surprisingly}
Vid Kocijan, Ana-Maria Cretu, Oana-Maria Camburu, Yordan Yordanov, and Thomas
  Lukasiewicz. 2019.
\newblock A surprisingly robust trick for winograd schema challenge.
\newblock \emph{arXiv preprint arXiv:1905.06290}.

\bibitem[{Levesque et~al.(2012)Levesque, Davis, and
  Morgenstern}]{levesque2012winograd}
Hector Levesque, Ernest Davis, and Leora Morgenstern. 2012.
\newblock The winograd schema challenge.
\newblock In \emph{Thirteenth International Conference on the Principles of
  Knowledge Representation and Reasoning}.

\bibitem[{Liu et~al.(2016)Liu, Lowe, Serban, Noseworthy, Charlin, and
  Pineau}]{liu2016not}
Chia-Wei Liu, Ryan Lowe, Iulian~V Serban, Michael Noseworthy, Laurent Charlin,
  and Joelle Pineau. 2016.
\newblock How not to evaluate your dialogue system: An empirical study of
  unsupervised evaluation metrics for dialogue response generation.
\newblock \emph{arXiv preprint arXiv:1603.08023}.

\bibitem[{Lowe et~al.(2017)Lowe, Noseworthy, Serban, Angelard-Gontier, Bengio,
  and Pineau}]{lowe-etal-2017-towards}
Ryan Lowe, Michael Noseworthy, Iulian~V Serban, Nicolas Angelard-Gontier,
  Yoshua Bengio, and Joelle Pineau. 2017.
\newblock Towards an automatic turing test: Learning to evaluate dialogue
  responses.
\newblock \emph{arXiv preprint arXiv:1708.07149}.

\bibitem[{Mehri and Eskenazi(2020)}]{mehri2020usr}
Shikib Mehri and Maxine Eskenazi. 2020.
\newblock Usr: An unsupervised and reference free evaluation metric for dialog
  generation.
\newblock \emph{arXiv preprint arXiv:2005.00456}.

\bibitem[{Qin et~al.(2019)Qin, Liu, Che, Wen, Li, and Liu}]{qin2019entity}
Libo Qin, Yijia Liu, Wanxiang Che, Haoyang Wen, Yangming Li, and Ting Liu.
  2019.
\newblock Entity-consistent end-to-end task-oriented dialogue system with kb
  retriever.
\newblock \emph{arXiv preprint arXiv:1909.06762}.

\bibitem[{Radford et~al.(2018)Radford, Narasimhan, Salimans, and
  Sutskever}]{radford2018improving}
Alec Radford, Karthik Narasimhan, Tim Salimans, and Ilya Sutskever. 2018.
\newblock Improving language understanding by generative pre-training.
\newblock \emph{URL https://s3-us-west-2. amazonaws.
  com/openai-assets/research-covers/languageunsupervised/language understanding
  paper. pdf}.

\bibitem[{Ram et~al.(2018)Ram, Prasad, Khatri, Venkatesh, Gabriel, Liu, Nunn,
  Hedayatnia, Cheng, Nagar et~al.}]{ram2018conversational}
Ashwin Ram, Rohit Prasad, Chandra Khatri, Anu Venkatesh, Raefer Gabriel, Qing
  Liu, Jeff Nunn, Behnam Hedayatnia, Ming Cheng, Ashish Nagar, et~al. 2018.
\newblock Conversational ai: The science behind the alexa prize.
\newblock \emph{arXiv preprint arXiv:1801.03604}.

\bibitem[{See et~al.(2019)See, Roller, Kiela, and Weston}]{see2019makes}
Abigail See, Stephen Roller, Douwe Kiela, and Jason Weston. 2019.
\newblock What makes a good conversation? how controllable attributes affect
  human judgments.
\newblock \emph{arXiv preprint arXiv:1902.08654}.

\bibitem[{Shin et~al.(2019)Shin, Xu, Madotto, and Fung}]{shin2019happybot}
Jamin Shin, Peng Xu, Andrea Madotto, and Pascale Fung. 2019.
\newblock Happybot: Generating empathetic dialogue responses by improving user
  experience look-ahead.
\newblock \emph{arXiv preprint arXiv:1906.08487}.

\bibitem[{Sordoni et~al.(2015)Sordoni, Galley, Auli, Brockett, Ji, Mitchell,
  Nie, Gao, and Dolan}]{Sordoni2015ANN}
Alessandro Sordoni, Michel Galley, Michael Auli, Chris Brockett, Yangfeng Ji,
  Margaret Mitchell, Jian-Yun Nie, Jianfeng Gao, and William~B. Dolan. 2015.
\newblock A neural network approach to context-sensitive generation of
  conversational responses.
\newblock In \emph{HLT-NAACL}.

\bibitem[{Venkatesh et~al.(2018)Venkatesh, Khatri, Ram, Guo, Gabriel, Nagar,
  Prasad, Cheng, Hedayatnia, Metallinou et~al.}]{venkatesh2018evaluating}
Anu Venkatesh, Chandra Khatri, Ashwin Ram, Fenfei Guo, Raefer Gabriel, Ashish
  Nagar, Rohit Prasad, Ming Cheng, Behnam Hedayatnia, Angeliki Metallinou,
  et~al. 2018.
\newblock On evaluating and comparing open domain dialog systems.
\newblock \emph{arXiv preprint arXiv:1801.03625}.

\bibitem[{Walker et~al.(1997)Walker, Litman, Kamm, and
  Abella}]{walker1997paradise}
Marilyn~A Walker, Diane~J Litman, Candace~A Kamm, and Alicia Abella. 1997.
\newblock Paradise: A framework for evaluating spoken dialogue agents.
\newblock \emph{arXiv preprint cmp-lg/9704004}.

\bibitem[{Yi et~al.(2019)Yi, Goel, Khatri, Chung, Hedayatnia, Venkatesh,
  Gabriel, and Hakkani-Tur}]{yi2019towards}
Sanghyun Yi, Rahul Goel, Chandra Khatri, Tagyoung Chung, Behnam Hedayatnia, Anu
  Venkatesh, Raefer Gabriel, and Dilek Hakkani-Tur. 2019.
\newblock Towards coherent and engaging spoken dialog response generation using
  automatic conversation evaluators.
\newblock \emph{arXiv preprint arXiv:1904.13015}.

\bibitem[{Zhang et~al.(2019)Zhang, Sun, Galley, Chen, Brockett, Gao, Gao, Liu,
  and Dolan}]{zhang2019dialogpt}
Yizhe Zhang, Siqi Sun, Michel Galley, Yen-Chun Chen, Chris Brockett, Xiang Gao,
  Jianfeng Gao, Jingjing Liu, and Bill Dolan. 2019.
\newblock Dialogpt: Large-scale generative pre-training for conversational
  response generation.
\newblock \emph{arXiv preprint arXiv:1911.00536}.

\bibitem[{Zhao et~al.(2017)Zhao, Zhao, and Eskenazi}]{zhao2017learning}
Tiancheng Zhao, Ran Zhao, and Maxine Eskenazi. 2017.
\newblock Learning discourse-level diversity for neural dialog models using
  conditional variational autoencoders.
\newblock \emph{arXiv preprint arXiv:1703.10960}.

\end{thebibliography}
\bibliographystyle{acl_natbib}

\end{document}